\documentclass{article}
\usepackage[dblblindworkshop, final]{neurips_2026}
\workshoptitle{Transitioning from Pre-Training to Post-Training}
\setcitestyle{authoryear,round,comma,sort&compress}
\usepackage[utf8]{inputenc} %
\usepackage[T1]{fontenc}    %
\usepackage{hyperref}       %
\hypersetup{hidelinks}
\usepackage{url}            %
\usepackage{booktabs}       %
\usepackage{tabularx}       %
\usepackage{graphicx}       %
\usepackage{flafter}        %
\usepackage{wrapfig}        %
\usepackage{amsfonts}       %
\usepackage{nicefrac}       %
\usepackage{microtype}      %
\usepackage{xcolor}         %
\usepackage{colortbl}       %

\title{Unlearnable, or Unmeasured? On the Reliability of Difficulty Labels in RLVR}

\author{%
  Chandak Chakma\quad
  Syed Nazmus Sakib\quad
  Nafiul Haque\quad
  Shifat E. Arman\thanks{Corresponding author: \texttt{shifatearman@du.ac.bd}}\\[4pt]
  \normalfont
  Department of Robotics and Mechatronics Engineering, University of Dhaka
}

\begin{document}

\raggedbottom

\maketitle

\begin{abstract}

  Reinforcement learning with verifiable rewards (RLVR) has become an important approach for improving reasoning during post-training. Recent work suggests that some difficult prompts remain resistant to learning even when they occasionally produce correct solutions. We revisit this unlearnability phenomenon and find that the affected prompts do improve, at roughly one third of the learnable rate, while the difficulty-defined set used to study them is much less reproducible than expected. These difficulty labels are estimated from a limited number of sampled responses. Combining them across seeds can further change which prompts are selected instead of simply reducing measurement noise. We develop a sampling-based framework for quantifying this instability and determining how much evaluation is required for difficulty assignments to reproduce reliably. We also revisit the gradient-similarity evidence proposed to explain unlearnability and show that part of the observed separation arises because difficult prompts provide fewer correct rollouts from which their gradients can be estimated. Matching this sample count weakens the gradient difference but does not remove it. Overall, the slow-learning phenomenon survives our reanalysis, while both the prompts used to define it and the evidence used to explain it require more careful measurement.
\end{abstract}

\section{Introduction}

Reinforcement learning with verifiable rewards (RLVR) has become one of the most effective approaches for improving reasoning in post-training. RLVR has produced substantial gains in mathematical and programmatic reasoning and now forms a central component of post-training for modern reasoning models \citep{lambert2024tulu,shao2024deepseekmath,guo2025deepseek,NEURIPS2025_a4277440}. At the same time, its success has exposed a more fundamental question: what determines whether a reasoning problem can actually be learned through RLVR? Some studies suggest that RLVR can push models toward stronger reasoning behavior, while others find that its gains remain closely tied to capabilities already present in the base policy \citep{NEURIPS2025_537d5aa7,wen2026reinforcement}. Understanding where learning succeeds, where it slows down, and why has therefore become a central problem in post-training.

Prompt difficulty has emerged as one way to study and control this phenomenon. Recent methods use a model's empirical success rate to decide which prompts should be replayed, filtered, routed, reweighted, or given additional training effort \citep{baroian2026prompt,luo2026drift,li2026enhancing,pang2026beyond,qu2026small,bae2026online,tang2026towards}. Difficulty is also used to probe the limits of RLVR itself. Most strikingly, \citet{chen2026unlearnability} identified a group of hard prompts that occasionally produce successful trajectories yet improve very little during training. They describe these prompts as \emph{unlearnable} and connect their behavior to lower within-group gradient similarity, suggesting a possible representation failure. If correct trajectories are already available but the model still fails to learn from them, the limitation would be deeper than ordinary exploration or reward sparsity.

Both difficulty-aware training and claims about unlearnability depend on how reliably hard prompts can be identified. Difficulty is estimated from sampled responses, so the same prompt can cross a threshold across repeated evaluations even when the model is unchanged. Aggregating these noisy labels can further change which prompts are selected rather than simply reduce measurement noise. We revisit the unlearnability result of \citet{chen2026unlearnability} from this perspective. The reported prompts still improve much more slowly than the learnable group, at roughly one third of the rate, but the set itself is highly sensitive to construction. Using the same candidate prompts, the published five-seed rule returns 74 prompts, while an alternative aggregation returns 228; individual seeds remain between 145 and 150. A same-checkpoint evaluation reproduces much of this disagreement without any change in training. The slow-learning separation therefore appears in our data, but the prompts used to define it are much less stable than the label suggests.

This raises a broader question: how much evaluation is needed for a difficulty label to be reliable? We show that reproducibility depends on the rollout budget and on how many prompts lie near the threshold. In our setting, 8 to 16 rollouts per prompt yield only 0.63 to 0.73 agreement, while 0.90 agreement requires a projected 242 rollouts. We also revisit the gradient-similarity explanation and find a separate measurement issue: hard prompts provide fewer correct rollouts, so their gradients are estimated from fewer samples. Matching this sample count reduces the easy-to-unlearnable similarity ratio from 2.327 times to 1.539 times, although a clear gap remains. Overall, the slow-learning phenomenon persists, but both the prompts used to define it and the evidence used to explain it depend strongly on measurement choices.

Our work makes three main contributions:

\begin{enumerate}

  \item \textbf{We revisit what ``unlearnable'' means in RLVR.}
  The reported prompts do improve during training, but at roughly one third of
  the learnable rate, supporting a persistent slow-learning effect rather than
  literal non-learning. At the same time, we show that the set used to define
  this effect is highly sensitive to its construction: changing the aggregation
  rule moves the set from 74 to 228 prompts, while much of the apparent
  disagreement across training seeds can be reproduced by resampling the same
  checkpoint.

  \item \textbf{We turn difficulty assignment into an explicit sampling-budget
  problem.} We characterize how reliably threshold-based difficulty labels can
  be reproduced as the number of rollouts increases, and use this relationship
  to determine the evaluation budget required for a chosen level of agreement.
  The resulting predictions match retrospective measurements and two
  prospectively frozen evaluations, providing a practical way to choose rollout
  budgets rather than setting them heuristically.

  \item \textbf{We reassess the proposed gradient explanation for
  unlearnability.} Hard prompts provide fewer correct rollouts, which makes
  their prompt-level gradients less precisely estimated. Matching the number of
  correct rollouts reduces the easy-to-unlearnable gradient-similarity ratio
  from 2.327 times to 1.539 times. A substantial difference remains, so the
  original evidence is weakened rather than eliminated, and the mechanism
  behind the slow-learning effect remains unresolved.

\end{enumerate}

\section{Related work}

Our work connects four lines of research in RLVR: understanding the limits of post-training, using prompt difficulty to guide training, explaining why some hard prompts remain difficult to learn, and studying the reliability of the measurements used to support these conclusions. We organize the related work around these four themes.

\paragraph{RLVR and the limits of post-training.}
Reinforcement learning with verifiable rewards has become an important approach
for improving reasoning after pre-training, building on the use of automatically
checkable answers to supervise mathematical reasoning \citep{cobbe2021training}. T\"ulu 3 established a practical
open recipe for reinforcement learning with automatically verifiable rewards,
while DeepSeek-R1 demonstrated that large-scale reinforcement learning can
substantially improve reasoning behavior \citep{lambert2024tulu,guo2025deepseek}. Recent systems such as DAPO have further studied how sampling and
optimization choices affect RLVR training \citep{NEURIPS2025_a4277440}. At the same time,
several works have questioned what this training can actually teach. \citet{NEURIPS2025_537d5aa7} argue that RLVR may primarily sharpen capabilities already present in
the base model rather than create entirely new reasoning capacity. Other work
reports that prolonged reinforcement learning can expand the reasoning boundary
beyond what is observed under shorter post-training runs \citep{liu2026prorl}.
\citet{chen2026unlearnability} identify a related phenomenon at the prompt level: some difficult
examples occasionally produce positive reward but improve very little during
training. They describe these prompts as \emph{unlearnable} and connect their
behavior to differences in gradient similarity. Our work starts from this
prompt-level observation, but asks whether the set of prompts used to support such
a trainability claim is itself measured reliably.

\paragraph{Difficulty as a training signal.}
Prompt difficulty is increasingly treated as an operational quantity rather than only a descriptive property. Recent methods use estimated success rates to decide which prompts should be replayed, routed, reweighted, or prioritized. Prompt Replay focuses training on high-signal prompts using on-policy reuse \citep{baroian2026prompt}. DRIFT routes examples according to difficulty and training dynamics \citep{luo2026drift}, while difficulty-aware group normalization
changes the treatment of examples from different difficulty bands \citep{li2026enhancing}. Other work uses rare-event statistics to allocate training effort \citep{pang2026beyond}, or predicts prompt difficulty with a smaller model before
expensive sampling is performed \citep{qu2026small}. Off-context GRPO instead
provides privileged information when hard prompts fail to produce useful
rollouts \citep{agrawal2026off}. Recent methods also use curriculum
construction and explicitly difficulty-aware optimization to decide which prompts
receive training effort \citep{gao2026prompt,dai2026harder}. Related methods use
risk-sensitive exploration or influence-based data selection to improve the
efficiency of RLVR \citep{jiang2026risk,zhu2026data}. Recent extreme
data-efficiency results further show that post-training outcomes can depend
strongly on which training examples are selected \citep{NEURIPS2025_b1ea3f93}.
These methods differ substantially in their
objectives, but they share an important dependency: prompt difficulty is
inferred from a finite number of sampled outcomes. We study the statistical
reliability of that underlying measurement.

\paragraph{Why hard prompts fail to learn.}
A second line of work asks how difficult examples can be made learnable and what
distinguishes successful from unsuccessful training signal. Process supervision
evaluates intermediate reasoning rather than only the final answer \citep{uesato2022solving,lightman2024let}, and recent RLVR methods combine process and
outcome information during training \citep{ye2025beyond}. Other methods provide
partial solution prefixes to increase the probability that hard prompts produce
informative trajectories \citep{beliaev2026max}. \citet{chen2026unlearnability} instead study the
geometry of per-prompt gradients and report lower gradient alignment for their
unlearnable set, interpreting this as evidence of a representation failure.
Gradient-based analyses have a broader history as tools for studying the
influence of individual training examples and optimization behavior \citep{koh2017understanding,pruthi2020estimating}. Recent work has also studied
optimization bias, entropy dynamics, token-level learning signal, credit
assignment, and gradient utilization as possible determinants of RLVR learning
behavior \citep{liu2025understanding,cui2025entropy,NEURIPS2025_a797c2d2,hao2026rethinking,pmlr-v267-kazemnejad25a,fu2026maspo}. We do not propose a new mechanism for
hard-prompt learning. Instead, we show that the gradient comparison itself
depends strongly on how many successful rollouts are averaged for each prompt.
Matching this sample count substantially reduces the reported separation, while
leaving a residual difference that still requires explanation.

\paragraph{Measurement and reproducibility.}
Reproducibility work in reinforcement learning and machine learning has shown
that seeds, evaluation protocols, benchmark variance, and search budgets can
materially change reported conclusions \citep{henderson2018deep,dodge2019show,agarwal2021deep,bouthillier2021accounting}. Our setting differs in
one important respect. We are not primarily asking whether an aggregate
benchmark score reproduces. We study whether the \emph{membership of a
prompt-level set} reproduces when that membership is defined by thresholding a
sampled success rate. This distinction matters because difficulty-aware
post-training methods increasingly act on such sets directly. To our knowledge,
prior RLVR work has not treated the aggregation rule and rollout budget behind
difficulty assignment as part of the measurement procedure whose reliability
must itself be established. Our
contribution is therefore complementary to both post-training research and the
broader reproducibility literature: we characterize when a prompt-level
trainability label is statistically stable enough to support downstream
scientific claims.

\section{Experimental Results}
\label{sec:experiments}

\subsection{Experimental setup}
\label{sec:setup}

We study the unlearnability phenomenon reported by \citet{chen2026unlearnability}. Our primary experiments use Qwen2.5-0.5B \citep{yang2024qwen25} trained with GRPO \citep{shao2024deepseekmath} on 1,023 prompts from MATH \citep{hendrycks2021measuring}. We train five independent seeds with \(G=8\) rollouts per prompt and use the original verifier without modification. We retain the published easy, learnable, and unlearnable cohorts and the original difficulty threshold \(\tau=0.1\). We write \(D_u\) for the population produced by the published unlearnability construction.

The experiments below separate three sources of variation that are easy to conflate: variation from training, variation from finite evaluation, and variation introduced by the rule used to aggregate prompt-level decisions. Unless stated otherwise, set-membership analyses use \(N=128\) evaluation rollouts per prompt and compare independently constructed sets using Jaccard similarity. Gradient analyses use \(N=200\) initial-policy rollouts and form each prompt-level gradient by averaging over its correct rollouts. Confidence intervals are obtained by bootstrap with prompts resampled at the level appropriate to each statistic.

Our implementation is not byte-for-byte identical to the original training configuration. The most important difference is generation length: our training and primary evaluation use a 1,024-token cap, whereas the original scripts default to 5,120. This matters more for hard prompts than for easy ones, so we test evaluation-side sensitivity to the longer cap directly and avoid transferring training-dependent magnitudes beyond our configuration. Appendix~\ref{app:replication-audit} reports the full constant-by-constant audit and all known differences. Upon acceptance, we will release code, rollout artifacts, the exact aggregation rules, and timestamped records of the two frozen predictions.

\subsection{The slow-learning separation appears, but literal non-learning does not}
\label{sec:slow_learning}

\begin{wrapfigure}{r}{0.48\textwidth}
    \vspace{-10pt}
    \centering
    \includegraphics[width=\linewidth]{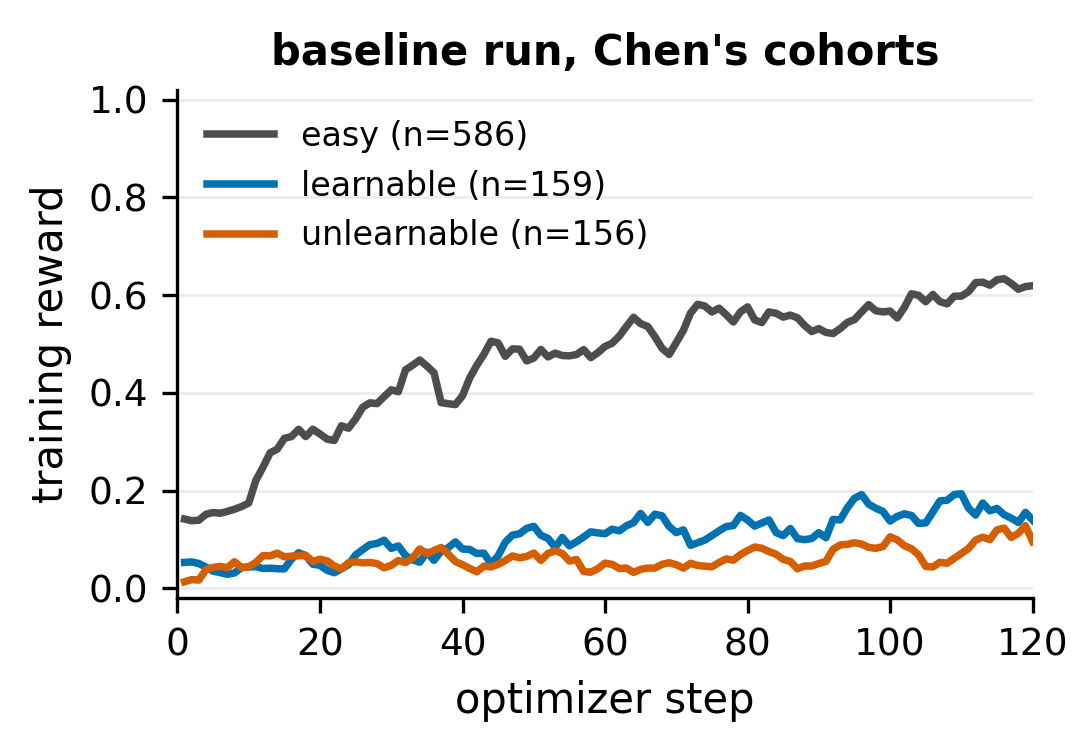}
    \caption{\textbf{The separation is present; the flatness is not.}
    Training reward for the published easy, learnable, and unlearnable cohorts.
    Slopes per 100 optimizer steps are \(+0.3800\), \(+0.1119\), and
    \(+0.0422\,[+0.0259,+0.0590]\), respectively. The affected cohort remains
    markedly slower to improve, but its slope is not zero.}
    \label{fig:slow_learning}
    \vspace{-8pt}
\end{wrapfigure}

\paragraph{Do the prompts labeled unlearnable actually fail to learn?}

We first ask whether the reported unlearnable prompts actually show no improvement during RLVR training. The three published groups remain clearly separated throughout training (Figure~\ref{fig:slow_learning}), but the \(D_u\) group is not flat. Reward increases by \(+0.0422\,[+0.0259,+0.0590]\) per 100 optimizer steps, compared with \(+0.1119\) for the learnable group and \(+0.3800\) for the easy group. Thus, the prompts labeled unlearnable do learn under RLVR, although much more slowly: their improvement is roughly one third of the learnable rate, and they remain at a substantially lower reward level throughout training.

We use the training slope rather than the terminal reward to characterize this difference. Under the GRPO sampling procedure, prompts stop contributing once the rewards within a sampled group have zero variance. As a result, the prompts still represented near the end of training are not necessarily the same prompts represented earlier, making the terminal cohort mean difficult to interpret as continued learning progress. The more defensible result is therefore a persistent difference in \emph{trainability}, rather than literal non-learning. We retain the notation \(D_u\) to remain consistent with the set defined by \citet{chen2026unlearnability}, but throughout this paper ``unlearnable'' refers to that published label, not to a claim that these prompts cannot improve.

\subsection{The published construction is not a consistent estimator of a fixed-threshold set}
\label{sec:set_instability}

\begin{figure}[t]
    \centering
    \includegraphics[width=0.98\linewidth]{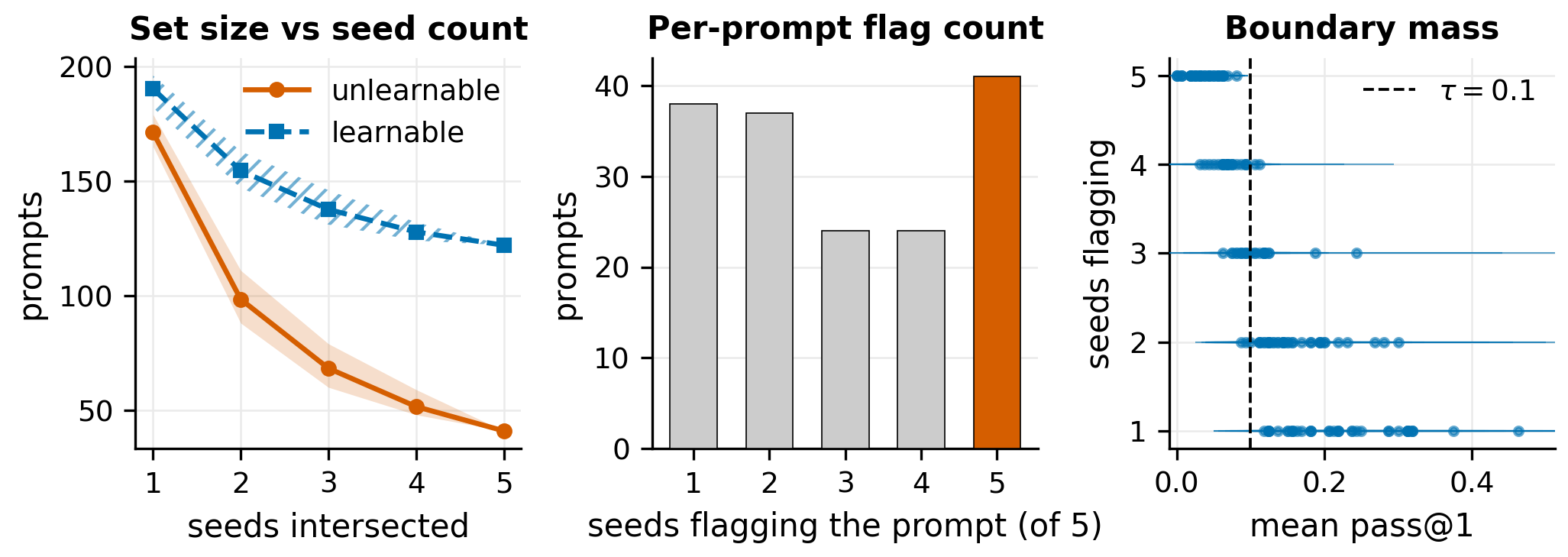}
    \vspace{-1.5mm}
    \includegraphics[width=0.88\linewidth]{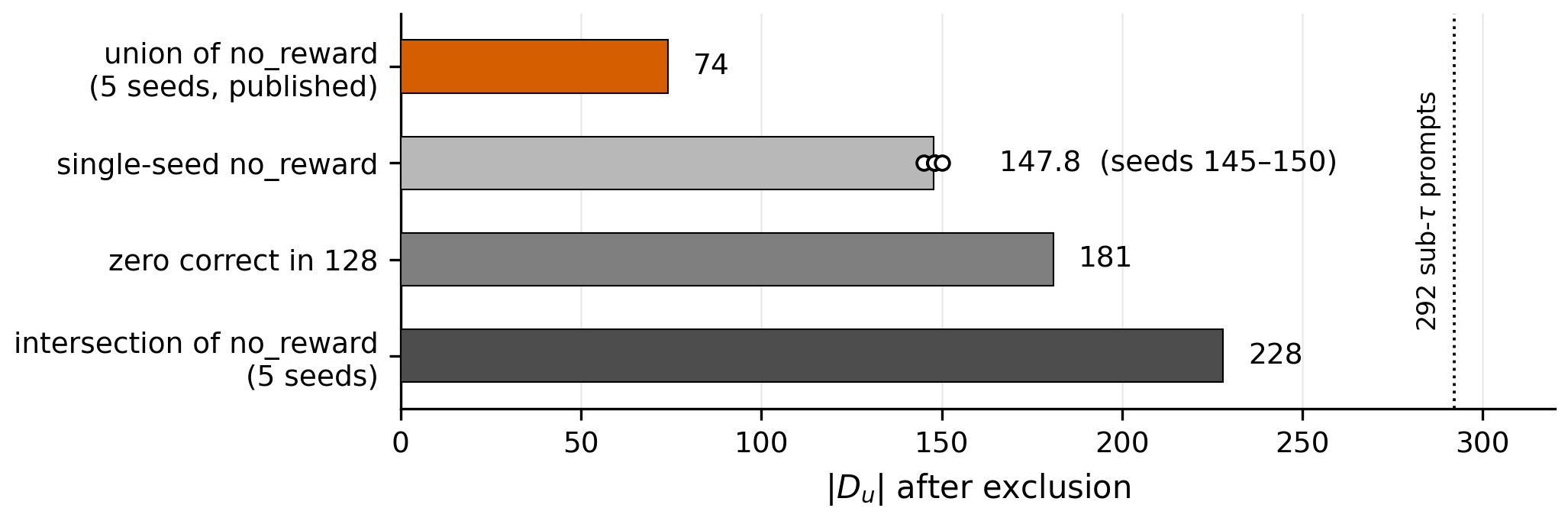}
    \caption{\textbf{The set is unstable in both seed count and aggregation operator.}
    Top: set size changes as more seeds are combined; many prompts are flagged by
    only a subset of seeds, with disagreement concentrated near \(\tau=0.1\).
    Bottom: holding the same 292 sub-threshold candidates fixed, the published
    five-seed construction returns 74 prompts, individual seeds return 145--150
    (mean 147.8), a zero-correct-in-128 rule returns 181, and the alternative
    five-seed aggregation returns 228.}
    \label{fig:set_construction}
\end{figure}

\paragraph{Does the aggregation rule matter more than the training seed?}

The previous result shows that the prompts labeled unlearnable are not literally static during training; they improve, but much more slowly than the learnable group. This makes the next question important: if we want to explain that slower learning by studying a particular set of prompts, how stable is that set itself? We therefore separate two possible sources of variation. One comes from training different seeds, while the other comes from the rule used to combine the seed-level labels. To compare them fairly, we first define a common group of candidate hard prompts. We evaluate each of the 1,023 prompts 128 times using the same reference checkpoint and compute its pass rate. Among them, 292 prompts have an estimated pass rate at or below the difficulty threshold \(\tau=0.1\), and we keep these 292 prompts fixed for the following comparison. Applying the published rule across all five training runs leaves only \textbf{74} prompts in \(D_u\), because a prompt is excluded if it is marked as \texttt{no\_reward} in any of the five runs. If we change only this aggregation step and exclude a prompt only when it is marked as \texttt{no\_reward} in all five runs, \textbf{228} prompts remain. In comparison, constructing \(D_u\) from each training seed separately gives much more similar set sizes, ranging from 145 to 150 prompts, with a mean of 147.8. A separate rule based only on the 128-response evaluation, where a prompt is selected if it receives zero correct responses, identifies 181 prompts. The final size of \(D_u\) therefore changes far more because of how the five runs are combined than because of which training seed is used (Figure~\ref{fig:set_construction}).

Figure~\ref{fig:set_construction} also shows that this is not simply a disagreement between two particular constructions. As additional seeds are combined, the selected set changes systematically rather than settling toward a stable collection of prompts. Many prompts are flagged by some seeds but not others, and most of this disagreement is concentrated near the threshold \(\tau=0.1\), where only a small change in the number of successful responses can move a prompt from one side of the threshold to the other. The instability is therefore not spread uniformly across prompts; it is concentrated exactly where the difficulty decision is most sensitive to finite sampling. This suggests that combining more seeds is not simply reducing noise. The aggregation rule is also changing which prompts remain in the final set. That leads to the next question: how much of the disagreement we observe is actually caused by training different models, and how much would still appear if we evaluated the same model again with a different set of sampled responses?

\paragraph{What fails: the latent quantity, or the rule used to estimate it?}
The underlying notion of difficulty is not the problem. If \(p(x)\) denotes the
true probability that a fixed policy solves prompt \(x\), then
\[
  D_{\tau}=\{x:p(x)\leq\tau\}
\]
defines a clear target set: all prompts whose true success probability is at or
below the chosen threshold \(\tau\). The difficulty comes from how this set is
estimated in practice. The published procedure observes only a limited number
of responses for each prompt, uses those responses to decide whether the prompt
falls below the threshold in each run, and then combines these yes-or-no
decisions across multiple runs. Combining the decisions in this way does more
than reduce sampling noise; it gradually changes which prompts can remain in
the set. In particular, if a prompt must be classified as difficult in every
independent run, then as more runs are added, only prompts that are almost
guaranteed to receive that label can survive. Under the binomial threshold
model, this eventually pushes the retained set toward prompts with \(p(x)=0\).
The subsequent \texttt{no\_reward} exclusion then removes exactly these prompts,
because they never produce a rewarded response. The procedure therefore does
not approach the fixed set \(D_{\tau}\) as more runs are added. It approaches a
degenerate outcome created by the aggregation rule itself. This inconsistency concerns the regime in which each run supplies a finite, independently thresholded label and the number of such runs increases; it is not a claim that every procedure using repeated evaluation is inconsistent, since pooling the same rollouts into a single deeper estimate provides a direct finite-sample estimate of $D_\tau$ whose sampling error decreases with rollout budget. The exclusion
creates a second problem: receiving no reward in the observed training runs is
not the same as being unsolvable. Of the 218 prompts removed by the
\texttt{no\_reward} criterion, 116 are solved at least once in a later \(N=128\)
evaluation. We therefore distinguish between prompts for which reward was not
observed during training and prompts that the policy cannot solve, rather than
treating the two as equivalent.

\paragraph{Is the disagreement mainly training variation?}
To remove training variation entirely, we evaluate the same checkpoint twice and change only the sampled rollouts. At \(N=32\), the two evaluations of the same model have Jaccard \(0.798\), compared with \(0.751\) across independently trained seeds. Expressed as disagreement, resampling one fixed model reproduces approximately 81\% of the apparent cross-seed effect. The same qualitative result remains at \(N=128\). Different training runs matter, but they are not required to generate most of the membership instability.

\paragraph{Could our shorter generation cap be creating the set movement?}
We re-evaluate all 1,023 prompts at \(N=128\) with a 5,120-token cap, changing nothing else. The 1,024- and 5,120-token evaluations agree at Jaccard \(0.8984\,[0.8639,0.9310]\), while two independent passes at the same budget agree at \(0.8770\,[0.8397,0.9127]\). Truncation falls from 0.0961 to 0.0320, but the movement in set membership remains within the ordinary repeatability scale. This control supports the evaluation-side instability result; it does not recreate training under the original 5,120-token configuration.

\subsection{How much evaluation does a difficulty-defined set need?}
\label{sec:budget}

The previous section identifies two different problems: the published aggregation operator changes the set it selects, and even a well-defined fixed-\(\tau\) set is noisy when \(p(x)\) is estimated from few rollouts. We now address the constructive question: if we want to estimate \(D_{\tau}\), how much evaluation is required?

\begin{figure*}[ht]
    \centering
    \begin{minipage}[t]{0.58\linewidth}
        \vspace{0pt}
        \centering
        \includegraphics[height=0.25\textheight,keepaspectratio]{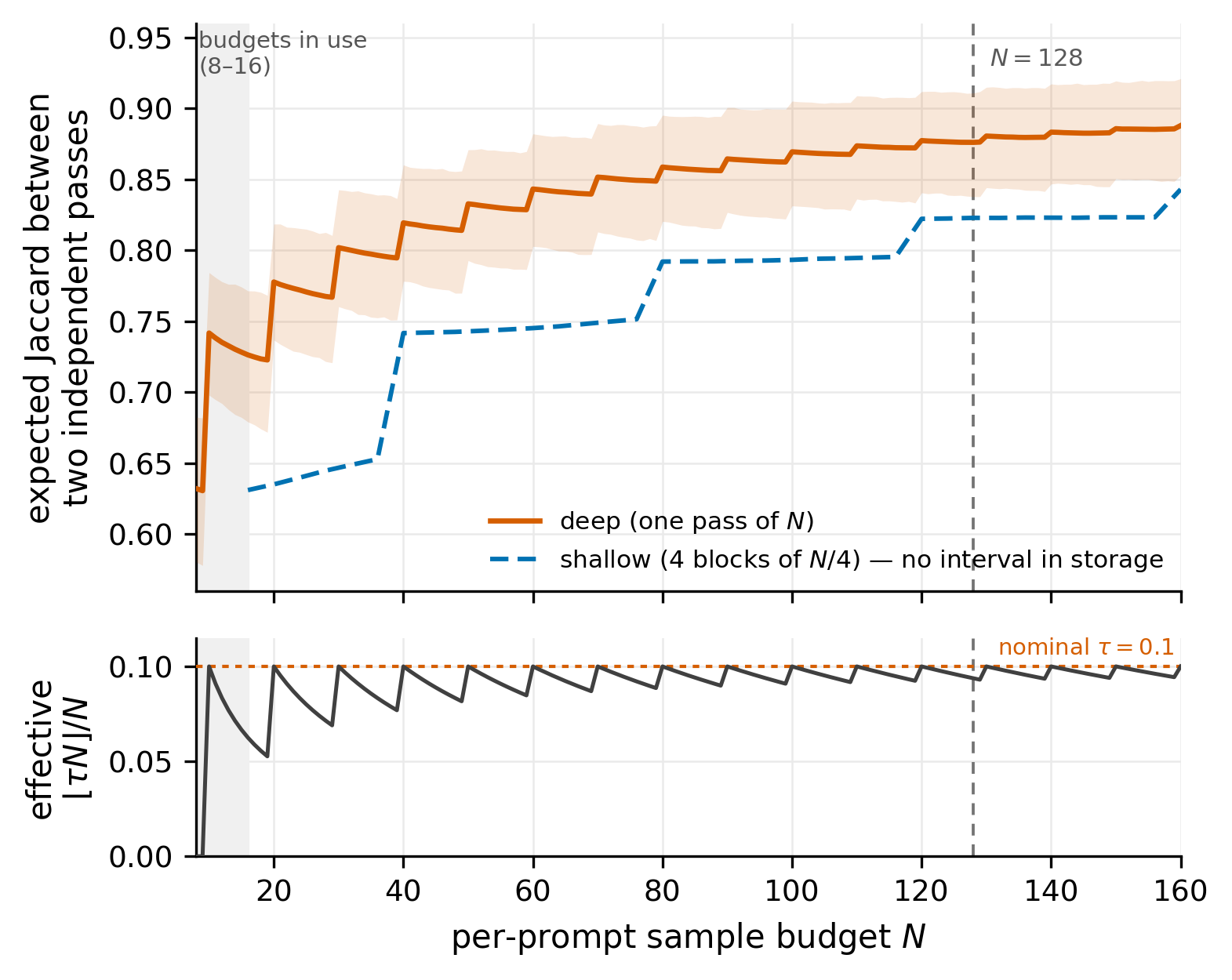}
        \par\vspace{-2mm}
        \small (a) Reproducibility as a function of sampling budget.
    \end{minipage}
    \hfill
    \begin{minipage}[t]{0.40\linewidth}
        \vspace{0pt}
        \centering
        \includegraphics[height=0.25\textheight,keepaspectratio]{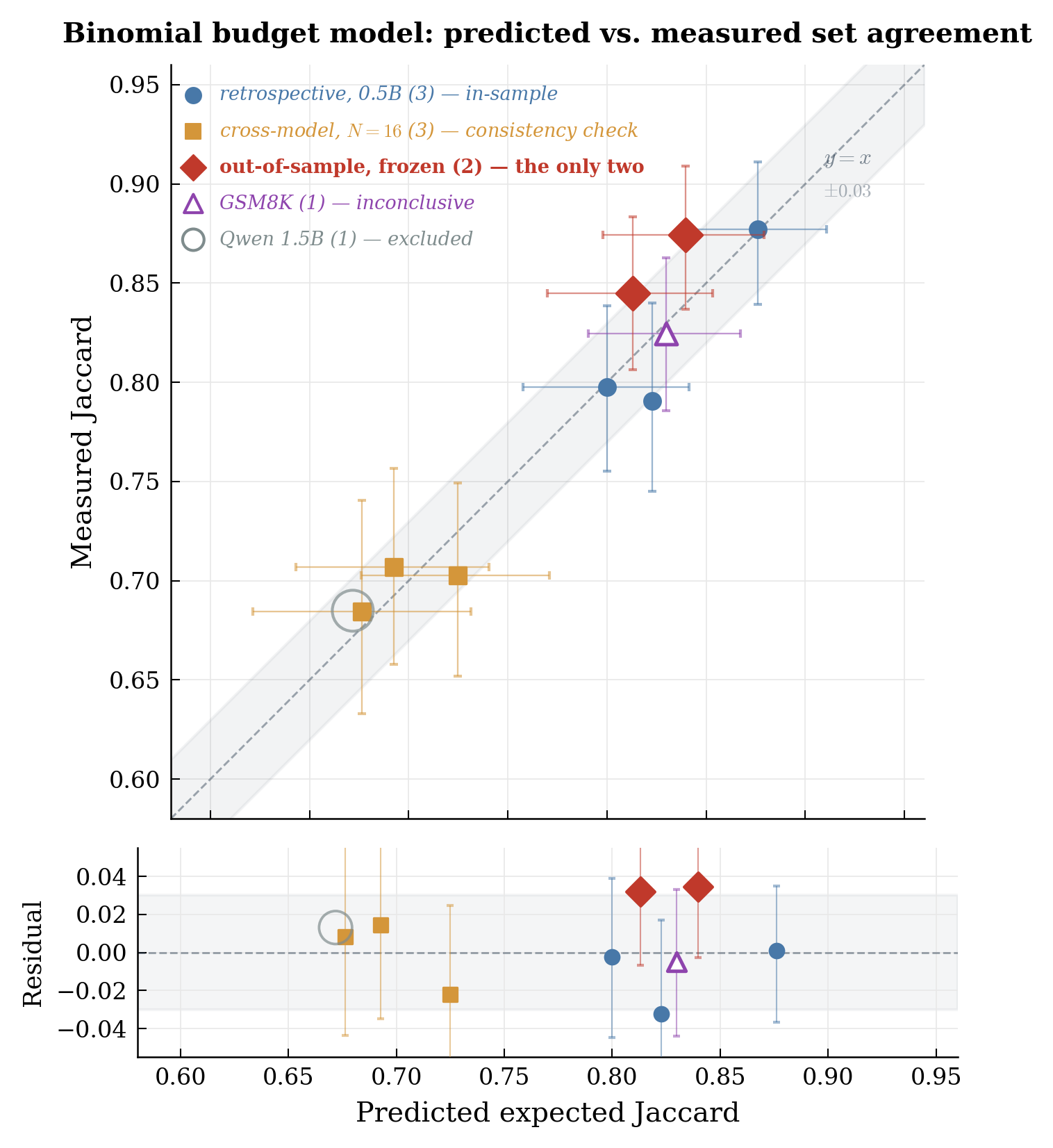}
        \par\vspace{-2mm}
        \small (b) Predicted versus measured agreement.
    \end{minipage}
    \caption{\textbf{The sampling budget can be predicted before it is spent.}
    (a) Predicted Jaccard agreement between two independent threshold-defined sets as a
    function of per-prompt budget \(N\); the lower panel shows the discrete
    effective threshold that produces the small-\(N\) sawtooth.
    (b) Predicted versus measured Jaccard. The filled diamonds are the two
    prospectively frozen \(N=40\) and \(N=64\) predictions. Retrospective and
    cross-model consistency checks are shown for context; the explicitly
    excluded 1.5B point is not treated as
    a validation result.}
    \label{fig:budget_validation}
\end{figure*}

\paragraph{At equal cost, should the budget be spent deeply or split into shallow passes?}
We compare two ways of using the same budget of 128 rollouts per prompt. In the
first, all 128 responses are used together to estimate the prompt's pass rate
and apply the difficulty threshold once. In the second, the 128 responses are
split into four independent blocks of 32, and a prompt is retained only if all
four blocks classify it as difficult. This gives a direct comparison because
both procedures apply the same effective threshold:
\[
  \frac{\lfloor 0.1\times128\rfloor}{128}
  =
  \frac{\lfloor 0.1\times32\rfloor}{32}
  =
  0.09375.
\]
Using all 128 rollouts together gives a Jaccard of
\(0.8770\,[0.8397,0.9127]\), while splitting them into four blocks gives
\(0.7906\,[0.7426,0.8398]\). The difference is
\(+0.0864\,[+0.0380,+0.1349]\). Since both procedures use exactly the same
number of rollouts, the improvement comes from how the budget is allocated
rather than from additional compute. This result is specific to evaluation on
a fixed checkpoint and does not imply that a single-pass rule is always optimal
when training variation is also involved.

\paragraph{How many rollouts are required for a target level of reproducibility?}
For a prompt with true success probability \(p\), evaluating it with \(N\)
rollouts gives a random number of correct responses, \(K\), following a binomial
distribution. A prompt is labeled difficult when the observed number of
successes falls below the threshold defined by \(\tau\). This gives \(q_N(p)\),
the probability that a prompt with success probability \(p\) receives the
difficult label at a particular rollout budget. To use this for the full set of
prompts, we need an estimate of how the underlying success probabilities are
distributed across the dataset. We obtain this from the pooled \(N=128\)
evaluation, while accounting for the fact that the observed pass rates are
themselves noisy finite-sample estimates. We do this using the
Kiefer--Wolfowitz nonparametric maximum-likelihood estimator
\citep{kieferWolfowitz1956}. Once this distribution is estimated, we can predict
how much two independent evaluations at the same rollout budget will agree with
each other. This gives the predicted Jaccard agreement for any choice of \(N\), and allows
us to determine how many rollouts are needed to reach a chosen level of
reproducibility.

Figure~\ref{fig:budget_validation}a shows how this agreement improves as the
evaluation budget increases. With only 8 rollouts per prompt, the predicted
Jaccard agreement is 0.6322; with 16 rollouts it rises to 0.7265, and with 128 rollouts it
reaches 0.8759. We can then ask for the rollout budget required to maintain a
particular level of agreement. In our setting, reaching a stable agreement of
0.80 requires about 40 rollouts per prompt,  while 0.90 and 0.95 require projected budgets of 242 and 2,111, both beyond our \(N=128\) fitting range. These values are specific to this model, dataset, and threshold,
because the required budget depends strongly on how many prompts lie close to
the decision boundary. The important contribution is therefore not the
particular number of rollouts, but the procedure for calculating it. The curve
also has a small sawtooth pattern at low budgets because the threshold can only
correspond to an integer number of successful responses. As \(N\) increases,
sampling noise decreases, but the effective cutoff occasionally shifts at the
same time. This means that adding a few more rollouts can sometimes reduce
agreement locally. For that reason, we report the first budget from which the
desired agreement remains satisfied, rather than an isolated point where the
curve happens to cross the target.

\paragraph{Does the budget model predict measurements that were not used to fit it?}
We first perform six retrospective checks, including three consistency checks on
additional models. We then freeze two predictions for budgets that had not yet
been measured and generate the corresponding evaluations only afterward.
For \(N=40\), the frozen prediction is
\(0.8131\,[0.7699,0.8533]\) and the measured Jaccard is
\(0.8450\,[0.8063,0.8833]\). For \(N=64\), the prediction is
\(0.8397\,[0.7980,0.8792]\) and the measurement is
\(0.8742\,[0.8369,0.9091]\). Both lie inside their pre-specified intervals
(Figure~\ref{fig:budget_validation}b). A homogeneous-pass-rate null can match
the marginal flagged fraction by construction but misses observed Jaccard by
0.19--0.53, whereas the distributional model misses the corresponding
retrospective measurements by only 0.008--0.032. Reproducibility is therefore
controlled by the distribution of prompt probabilities around the threshold,
not merely by the final set size.

\begin{figure*}[ht]
    \centering
    \includegraphics[width=0.96\linewidth]{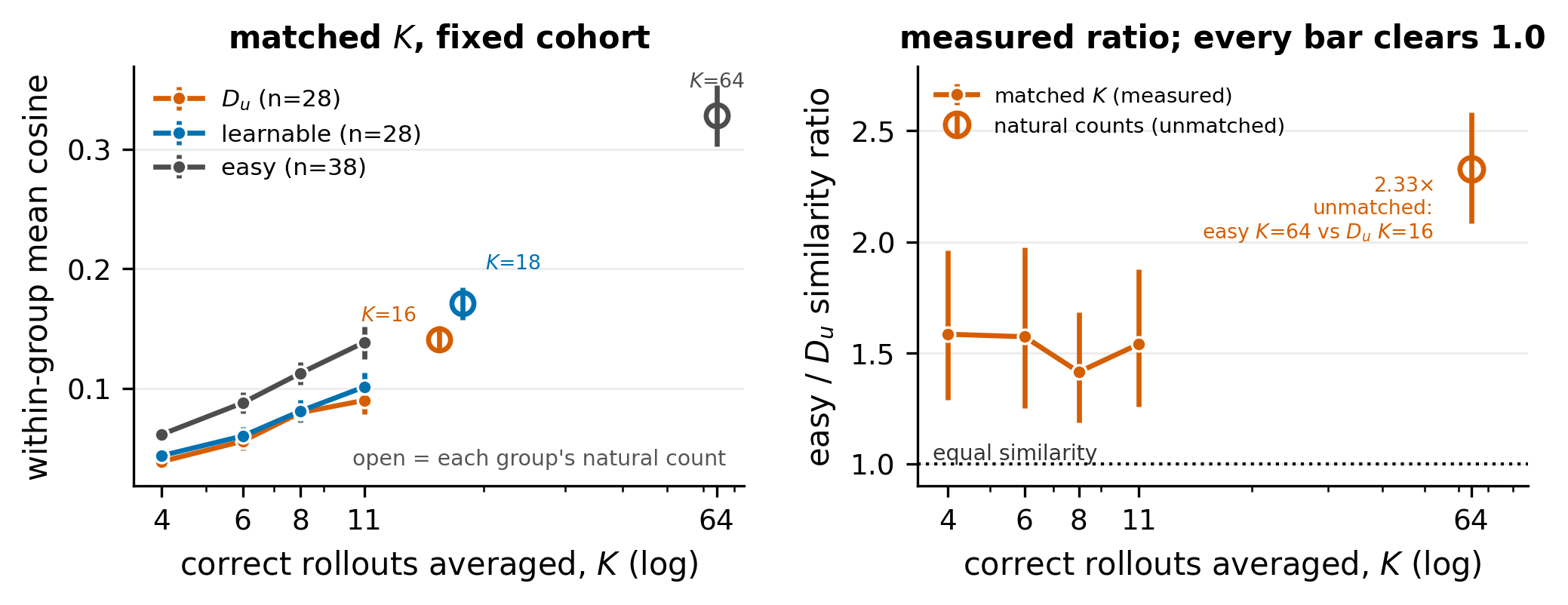}
    \caption{\textbf{Gradient similarity after matching estimator sample count.}
    Left: within-group mean cosine as the number \(K\) of correct rollouts
    averaged per prompt is matched across cohorts. The fixed matched cohort
    contains 28 \(D_u\), 28 learnable, and 38 easy prompts. Right: the easy-to-\(D_u\)
    similarity ratio. Matching reduces the unmatched 2.327 times ratio to
    approximately 1.5 times, while every tested matched level remains above
    one. Open markers show each group's natural, unmatched sample count. At $K=11$, matching retains only 28 of the 71 $D_u$ prompts in the gradient cohort; the matched values therefore demonstrate sensitivity to unequal estimator sample count in an eligible subset, rather than providing corrected full-cohort estimates.}
    \label{fig:matched_k}
\end{figure*}

\subsection{Does the gradient gap persist after matching sample count?}
\label{sec:gradient}

The proposed explanation for unlearnability relies partly on the observation
that gradients from \(D_u\) prompts are less similar to one another than
gradients from easy prompts. However, the two groups are not measured under the
same conditions. Each prompt-level gradient is estimated by averaging the
gradients of its correct rollouts, and difficult prompts naturally produce
fewer correct responses. In the \(N=200\) gradient cohort, \(D_u\) prompts have
a median of only 8 correct rollouts, compared with 65 for easy prompts. Their
gradient estimates are therefore based on substantially fewer samples,
creating a separate measurement problem from the instability of the difficulty
labels.

To test how much this difference in sample count affects the result, we
recompute the comparison using the same number of correct rollouts for every
prompt. In the original unmatched analysis, the easy-to-\(D_u\) similarity
ratio is 2.327 times. When both groups are restricted to \(K=11\) correct
rollouts per prompt, the ratio falls to 1.539 times, with a 95\% confidence
interval of \([1.26,1.88]\) (Figure~\ref{fig:matched_k}). The same pattern
appears at \(K=4,6,8,\) and \(11\): every matched estimate is smaller than the
original ratio, while all remain above one. Matching the sample count therefore
weakens the reported gradient separation substantially, but does not eliminate
it.

This comparison also introduces an important selection effect. Requiring
\(K=11\) correct rollouts leaves only 28 of the original 71 \(D_u\) prompts
eligible, together with 38 easy prompts. The retained \(D_u\) prompts are
necessarily those that produced more successful responses and are therefore
easier than many of the prompts removed by the matching requirement. On a log
scale, matching removes 48.9\% of the observed separation, with an interval of
\([29.3,71.2]\%\), but we do not interpret this as a causal decomposition. The
matched result demonstrates that unequal sample count contributes substantially
to the original gradient gap; it is not a corrected estimate for the full
\(D_u\) set. Full attrition and alternative matching levels are reported in Appendix~C.

We next consider whether the remaining slow-learning gap can be explained by
how often difficult prompts contribute to GRPO updates. With \(G=8\), a prompt
produces no update when all sampled rewards are identical, so difficult prompts
have fewer opportunities to contribute.
\(D_u\) prompts receive an average of
2.14 contributing steps per run, compared with 3.13 for learnable prompts and
5.26 for easy prompts. This difference is substantial, but it does not explain
the phenomenon on its own. None of the 156 prompts in the exposure-analysis
cohort is completely starved across all five runs, and 35 of them, or 22.4\%, receive at least the
mean exposure of the learnable group while still improving slowly.

These results rule out the simple form of exposure starvation in which \(D_u\)
prompts fail to learn because they receive no usable update opportunities. They
do not rule out a stronger possibility in which the rewarded trajectories
themselves contain weak or uninformative learning signal, since
outcome-equivalent trajectories can still differ in reasoning quality
\citep{mei2026good}. We cannot test that
explanation directly because the training-time response strings were not
stored. We therefore distinguish between \emph{exposure starvation}, which is
not supported by our measurements, and \emph{content starvation}, which remains
unresolved.

A second possibility is that GRPO's reward-standard-deviation normalization
systematically suppresses updates on hard prompts. If this were the case,
\(D_u\) prompts should receive smaller update scales on the steps where they
contribute. We observe the opposite. Conditional on a prompt producing
non-zero reward variance, the mean \(1/\sigma\) scale is 2.8360 for \(D_u\)
prompts and 2.4262 for easy prompts, about 17\% larger for the harder group. The
main asymmetry is therefore in how often a usable update is formed, not in a
smaller normalization factor once an update exists. Together, these results
substantially weaken the original gradient-based explanation and rule out two
simple optimization explanations in the forms we can directly measure, while
leaving the underlying cause of the slow-learning effect unresolved.

\subsection{An unstable difficulty rule does not necessarily imply unstable training}
\label{sec:training_effect}

The results above concern the reliability of a difficulty-defined set and the
scientific claims made from it, but a separate question is whether an unstable
difficulty rule necessarily makes difficulty-aware RLVR ineffective. We test
this using two neighboring constructions that remove very different amounts of
usable training signal: after accounting for GRPO's zero-variance filter, one
removes approximately 5.0\% of the usable training signal while the other
removes 27.6\%, a difference of more than fivefold. We train five seeds under
each construction and compare them with size-matched random-removal controls.
Despite this large difference in removed signal, the resulting models differ by
less than one held-out accuracy point. The registered sensitivity of the
experiment is not sufficient to resolve effects of that size, so this should
not be interpreted as evidence of equivalence. It does, however, establish an
important boundary on our claim: a large change in the reported set does not
produce an equally large change in training performance in this setting. Our strongest conclusion is therefore about measurement and inference: our experiments do not show that every training method using a noisy difficulty signal must fail.

\section{Conclusion}

We revisited the recently reported unlearnability phenomenon in RLVR and found a more nuanced result. The affected prompts improve at roughly one third of the learnable rate rather than not at all, but the prompts used to define them are far less stable than the effect itself. The published aggregation rule does not consistently recover a fixed-threshold set, and finite-sample evaluation explains much of the apparent disagreement across training seeds. We provide a way to determine how much evaluation is required for reproducible difficulty assignment and validate its predictions at previously unmeasured budgets. We also show that the gradient-similarity difference proposed to explain unlearnability shrinks substantially when estimator sample count is matched, although a residual gap remains. The main lesson is therefore not that unlearnability disappears, but that low-budget difficulty labels are unreliable for classifying individual prompts as "unlearnable," or for supporting mechanistic claims built on that classification.

\begingroup
\raggedright
\bibliographystyle{plainnat}
\bibliography{ref}
\endgroup

\clearpage
\phantomsection
\section*{Appendix --- Contents and Index}
\addcontentsline{toc}{section}{Appendix --- Contents and Index}
\label{sec:appendix-contents}

\noindent\textbf{Appendix A \quad Reproduction and set construction in full} \dotfill p.~\pageref{app:set-construction}\\
\hspace*{1.2em} A.1~~The slow-learning effect \dotfill p.~\pageref{app:slow-learning}\\
\hspace*{1.2em} A.2~~How the published construction changes the selected set \dotfill p.~\pageref{app:published-construction}\\
\hspace*{1.2em} A.3~~Evaluation noise versus training-seed variation \dotfill p.~\pageref{app:eval-noise}\par\vspace{0.4em}

\noindent\textbf{Appendix B \quad Sampling-budget model and validation} \dotfill p.~\pageref{app:sampling-budget}\\
\hspace*{1.2em} B.1~~From success probability to reproducibility \dotfill p.~\pageref{app:success-to-reproducibility}\\
\hspace*{1.2em} B.2~~Equal-cost allocation of the rollout budget \dotfill p.~\pageref{app:equal-cost}\\
\hspace*{1.2em} B.3~~Validation of the sampling-budget model \dotfill p.~\pageref{app:validation}\\
\hspace*{1.2em} B.4~~Discrete thresholds and what transfers across settings \dotfill p.~\pageref{app:discrete-thresholds}\par\vspace{0.4em}

\noindent\textbf{Appendix C \quad Gradient and downstream sensitivity analyses} \dotfill p.~\pageref{app:gradient-sensitivity}\\
\hspace*{1.2em} C.1~~Propagation of set uncertainty to gradient similarity \dotfill p.~\pageref{app:propagation}\\
\hspace*{1.2em} C.2~~Matched-gradient analysis and cohort composition \dotfill p.~\pageref{app:matched-gradient}\par\vspace{0.4em}

\noindent\textbf{Appendix D \quad Optimization diagnostics and training impact} \dotfill p.~\pageref{app:optimization-diagnostics}\\
\hspace*{1.2em} D.1~~Exposure, zero-variance filtering, and GRPO normalization \dotfill p.~\pageref{app:exposure-normalization}\\
\hspace*{1.2em} D.2~~Does difficulty construction materially change training? \dotfill p.~\pageref{app:training-impact}\par\vspace{0.4em}

\noindent\textbf{Appendix E \quad Experimental configuration and replication audit} \dotfill p.~\pageref{app:replication-audit}\\
\hspace*{1.2em} E.1~~Experimental configuration and statistical procedures \dotfill p.~\pageref{app:replication-audit-config}\\
\hspace*{1.2em} E.2~~Generation length and answer extraction \dotfill p.~\pageref{app:generation-length}\par

\clearpage
\appendix

\section{Reproduction and set construction in full}
\label{app:set-construction}

\subsection{The slow-learning effect}
\label{app:slow-learning}

We first reproduce the training behavior using the easy, learnable, and unlearnable groups published by \citet{chen2026unlearnability}. The groups remain clearly separated throughout training. Estimated reward slopes per 100 optimizer steps are
\[
+0.3800 \quad (\mathrm{easy}), \qquad
+0.1119 \quad (\mathrm{learnable}), \qquad
+0.0422\,[+0.0259,+0.0590] \quad (D_u).
\]

The \(D_u\) interval excludes zero. The affected prompts therefore improve during training, but substantially more slowly than the learnable group. At step 120, the difference between the learnable and \(D_u\) groups is \(+0.0754\,[+0.0279,+0.1246]\).

We rely on the slope rather than the terminal reward level. In this GRPO implementation, a prompt stops contributing once the sampled rewards have zero within-group variance. As training progresses, prompts that become consistently solved or consistently failed can therefore disappear from subsequent sampling. The terminal cohort mean is consequently affected by which prompts remain active. The slope gives the more defensible summary of whether the \(D_u\) group changes over training.

The reproduction supports a persistent difference in trainability, but not literal non-learning. Throughout the paper, \(D_u\) refers to the set produced by the published construction rather than to a claim that every member is intrinsically unlearnable.

\subsection{How the published construction changes the selected set}
\label{app:published-construction}

Let \(p(x)\) denote the true success probability of prompt \(x\) under a fixed policy. A fixed-threshold difficulty set,
\[
D_{\tau} = \{x:p(x)\leq\tau\},
\]
is well defined. The published construction does not observe \(p(x)\) directly. It first thresholds finite-sample success counts within individual runs and then combines those binary decisions across seeds. This distinction produces two separate sources of instability. First, prompts near \(\tau\) can cross the threshold under repeated sampling even when the policy does not change. Second, applying intersections and exclusions across these noisy binary labels changes the set being selected as more runs are added. Holding the same 292 prompts identified as sub-threshold in a pooled \(N=128\) evaluation fixed, the published five-seed construction returns 74 prompts. Individual seeds return between 145 and 150 prompts, with mean 147.8. A direct rule requiring zero successes in 128 evaluation samples returns 181, while the alternative five-seed aggregation returns 228. The difference caused by the aggregation operator is therefore much larger than the variation across individual training seeds.

The seed-count analysis shows the same behavior continuously rather than only at the two endpoint constructions. As more seeds are combined, the selected set shrinks instead of approaching a stable finite-\(\tau\) set. Many prompts are flagged by only a subset of the five runs, and this disagreement is concentrated near \(\tau=0.1\). Under repeated independent application, an intersection retains only prompts whose probability of being flagged approaches one. For the binomial threshold rule considered here, this pushes the limit toward prompts with \(p=0\). The accompanying never-rewarded exclusion then removes those prompts as the number of opportunities to observe the exclusion condition grows. The resulting procedure therefore has a degenerate limiting behavior rather than consistently estimating \(D_{\tau}\).

The same preference for the lower tail is visible in a direct precision--recall comparison. Against the \(N=128\) reference set, the five-seed intersection has precision \(0.976\) and recall \(0.541\,[0.421,0.657]\). The construction therefore does not simply recover a cleaner version of the threshold-defined set. It preferentially keeps prompts far enough below the threshold to survive repeated noisy decisions. This behavior is important when interpreting intersections as a denoising operation. High precision alone can make the resulting set appear more reliable, but the corresponding loss in recall shows that the rule has changed the selected region of the difficulty spectrum. It acts as a selector for the lower tail rather than as a neutral estimator of the original fixed threshold. For prompts near the decision boundary, a more natural representation is therefore uncertainty in membership rather than a deterministic label. The sampling-budget analysis in Appendix~\ref{app:sampling-budget} asks the related operational question: how many responses are required before that uncertainty becomes small enough for a binary label to reproduce reliably?

The published construction also removes prompts that never received positive reward during the training runs. In our data, this exclusion removes 218 prompts. A later \(N=128\) evaluation solves 116 of those prompts at least once. For this reason, we keep two constructions separate in the full analysis. The published observed-reward construction gives
\[
|D_u^{\mathrm{observed}}| = 74,
\]
whereas defining the exclusion using demonstrated solvability under the deeper evaluation gives
\[
|D_u^{\mathrm{solvable}}| = 181.
\]
The two sets overlap on 65 prompts. We use the 74-prompt set whenever reproducing the original claim, because that is the construction under audit, but we do not interpret the excluded prompts as demonstrated to be impossible for the model. The distinction is simple but important. ``No positive reward was observed during these training runs'' and ``the policy cannot solve this prompt'' are different statements. The former depends on the number of opportunities the prompt received and on the sampled trajectories; the latter is a property of the underlying policy and task. Together, these analyses separate three statements that would otherwise be easy to conflate: hard prompts exist, finite-sample estimates of difficulty can be unstable, and the particular multi-seed construction used to define \(D_u\) changes the set being selected. Our results challenge the last two without denying the first.

\subsection{Evaluation noise versus training-seed variation}
\label{app:eval-noise}

To determine how much of the disagreement actually comes from training, we repeat the difficulty assignment on the same fixed checkpoint and change only the sampled responses. At \(N=32\), two evaluations of the same checkpoint have Jaccard similarity \(0.798\). Across independently trained seeds, the corresponding similarity is \(0.751\). Expressed in terms of disagreement, the same-checkpoint experiment reproduces approximately 81\% of the apparent cross-seed effect. At \(N=128\), the disagreement is \(0.123\) for repeated evaluation of the same model and \(0.166\) across seeds, corresponding to a similar share of approximately 74\%. The \(N=32\) cross-seed quantity averages ten seed pairs, whereas the \(N=128\) result is based on a single pair, so we do not treat the two percentages as directly comparable estimates of one underlying constant.

A binomial model of the threshold decision predicts the same-model agreement to within \(0.004\). This supports the interpretation that finite evaluation sampling accounts for a large fraction of the observed membership instability.

Table~\ref{tab:robustness} summarizes the main controls used to separate evaluation noise, generation-length effects, and sampling-budget allocation. The corresponding analyses are reported in Appendix~A.3, Appendix~E.2, and Appendix~B.2, respectively.

\begin{table*}[ht]
\centering
\small
\setlength{\tabcolsep}{5pt}
\renewcommand{\arraystretch}{1.16}
\begin{tabularx}{\linewidth}{@{}
  >{\raggedright\arraybackslash}p{0.19\linewidth}
  >{\raggedright\arraybackslash}p{0.24\linewidth}
  >{\raggedright\arraybackslash}p{0.18\linewidth}
  >{\raggedright\arraybackslash}X@{}}
\toprule
\rowcolor{gray!14}
\textbf{Control} & \textbf{Comparison} & \textbf{Result} & \textbf{Interpretation} \\
\midrule
\rowcolor{gray!4}
Same-model resampling
& fixed checkpoint vs.\ cross-seed, \(N=32\)
& 0.798 vs.\ 0.751
& Most apparent cross-seed disagreement can arise from evaluation sampling alone. \\
Token-cap sensitivity
& 1,024 vs.\ 5,120 tokens
& 0.898 Jaccard
& Set movement is no larger than the measured repeatability scale. \\
\rowcolor{gray!4}
Equal-cost allocation
& one \(N=128\) pass vs.\ \(4\times32\)
& 0.877 vs.\ 0.791
& A deeper estimate is more reproducible at the same rollout cost. \\
\bottomrule
\end{tabularx}
\caption{\textbf{Key robustness controls.} The first control separates evaluation
noise from training variation; the second tests the generation-cap confound; the
third asks how an equal sampling budget should be allocated.}
\label{tab:controls}
\label{tab:robustness}
\end{table*}

\section{Sampling-budget model and validation}
\label{app:sampling-budget}

\subsection{From success probability to reproducibility}
\label{app:success-to-reproducibility}

The main paper treats difficulty assignment as a thresholding problem. For a prompt with latent success probability \(p\), an evaluation with \(N\) sampled responses produces
\[
K \mid p \sim \mathrm{Binomial}(N,p),
\]
and the prompt is assigned to the difficulty-defined set whenever
\[
K \leq \lfloor \tau N \rfloor.
\]

The corresponding probability that a prompt with success probability \(p\) is flagged is therefore
\[
q_N(p)
=
\Pr\!\left(K\leq \lfloor\tau N\rfloor \mid p\right).
\]

This quantity makes the source of instability explicit. Prompts far below the threshold have \(q_N(p)\) close to one, while prompts far above it have \(q_N(p)\) close to zero. The uncertainty is concentrated near the threshold, where the same unchanged prompt can be included in one evaluation and excluded in another. The reproducibility of the final set therefore depends not only on \(N\), but also on how much of the prompt distribution lies in this uncertain region.

We estimate that distribution from the pooled \(N=128\) evaluation. Directly using the empirical pass rates would treat their own sampling noise as genuine heterogeneity and would consequently overstate the stability of the thresholding procedure. We instead estimate a nonparametric mixing distribution over latent pass rates using the Kiefer--Wolfowitz maximum-likelihood formulation \citep{kieferWolfowitz1956}. The resulting distribution is used throughout the budget analysis; the raw empirical rates are not treated as noise-free estimates of \(p\).

For two independent evaluations of the same prompt, the probability that it is included in both sets is \(q_N(p)^2\), while the probability that it appears in their union is \(2q_N(p)-q_N(p)^2\). Integrating these quantities over the estimated pass-rate distribution gives the expected intersection and union rates for two independent applications of the threshold rule. Their ratio gives the predicted population-level Jaccard agreement used throughout the paper. This analytic quantity is the ratio of expected membership rates, not the expectation of the finite-set Jaccard ratio itself,
\[
J_N
=
\frac{
\int q_N(p)^2\,dF(p)
}{
\int \left[2q_N(p)-q_N(p)^2\right]\,dF(p)
},
\]
where \(F\) is the estimated distribution of latent prompt success probabilities. We evaluate this quantity over a dense integer grid of sampling budgets and invert the resulting curve to find the smallest budget from which a chosen reproducibility target remains satisfied.

The distinction between the first crossing and a stable crossing matters because the curve is not monotonic at small \(N\). The threshold is implemented through an integer number of successes, so the effective cutoff is
\[
\tau_{\mathrm{eff}}(N)
=
\frac{\lfloor\tau N\rfloor}{N},
\]
rather than exactly \(\tau\). Between successive values of \(N\) at which \(\lfloor\tau N\rfloor\) increases, the effective threshold moves downward. Increasing \(N\) can therefore reduce sampling variance while simultaneously making the decision boundary stricter. The sawtooth pattern in the main paper is a direct consequence of this interaction, not Monte Carlo noise.

For the Qwen2.5-0.5B/MATH distribution at \(\tau=0.1\), the predicted Jaccard agreement values at representative budgets are:
\[
\begin{array}{c|ccccccc}
N & 8 & 16 & 32 & 64 & 128 & 256 & 512 \\
\hline
\mathrm{Deep} &
0.6322 &
0.7265 &
0.7998 &
0.8413 &
0.8759 &
0.9022 &
0.9171
\end{array}
\]

A stable target of 0.80 requires approximately \(N=40\), 0.90 requires \(N=242\), and 0.95 requires \(N=2111\). These values are properties of this particular threshold and pass-rate distribution; they are not universal recommendations. The transferable object is the calculation itself.

\subsection{Equal-cost allocation of the rollout budget}
\label{app:equal-cost}

The main paper compares two ways of spending the same evaluation budget. The \textbf{deep} rule uses all 128 responses to form one pass-rate estimate. The \textbf{shallow} rule divides those 128 responses into four disjoint blocks of 32 and retains a prompt only if all four blocks flag it. Both procedures therefore see exactly the same number of responses per prompt.

This comparison is interpretable only because their effective thresholds happen to match:
\[
\frac{\lfloor 0.1\times128\rfloor}{128}
=
\frac{\lfloor 0.1\times32\rfloor}{32}
=
0.09375.
\]

This equality is specific to the chosen pair of budgets. In general, comparing one deep estimate with several shallow estimates can silently compare different effective thresholds, in which case any change in reproducibility mixes the effect of aggregation with a change in the underlying decision rule.

At equal cost, the deep rule is more reproducible. Its Jaccard across two independent evaluations is
\[
0.8770\,[0.8397,0.9127],
\]
compared with
\[
0.7906\,[0.7426,0.8398]
\]
for the four-block intersection. The paired difference is
\[
+0.0864\,[+0.0380,+0.1349].
\]

The same ordering remains after applying the never-rewarded exclusion. On the 759-prompt eligible subset, the difference grows to
\[
+0.2138\,[+0.0905,+0.3354].
\]

\begin{figure}[t]
  \centering
  \includegraphics[width=0.68\linewidth]{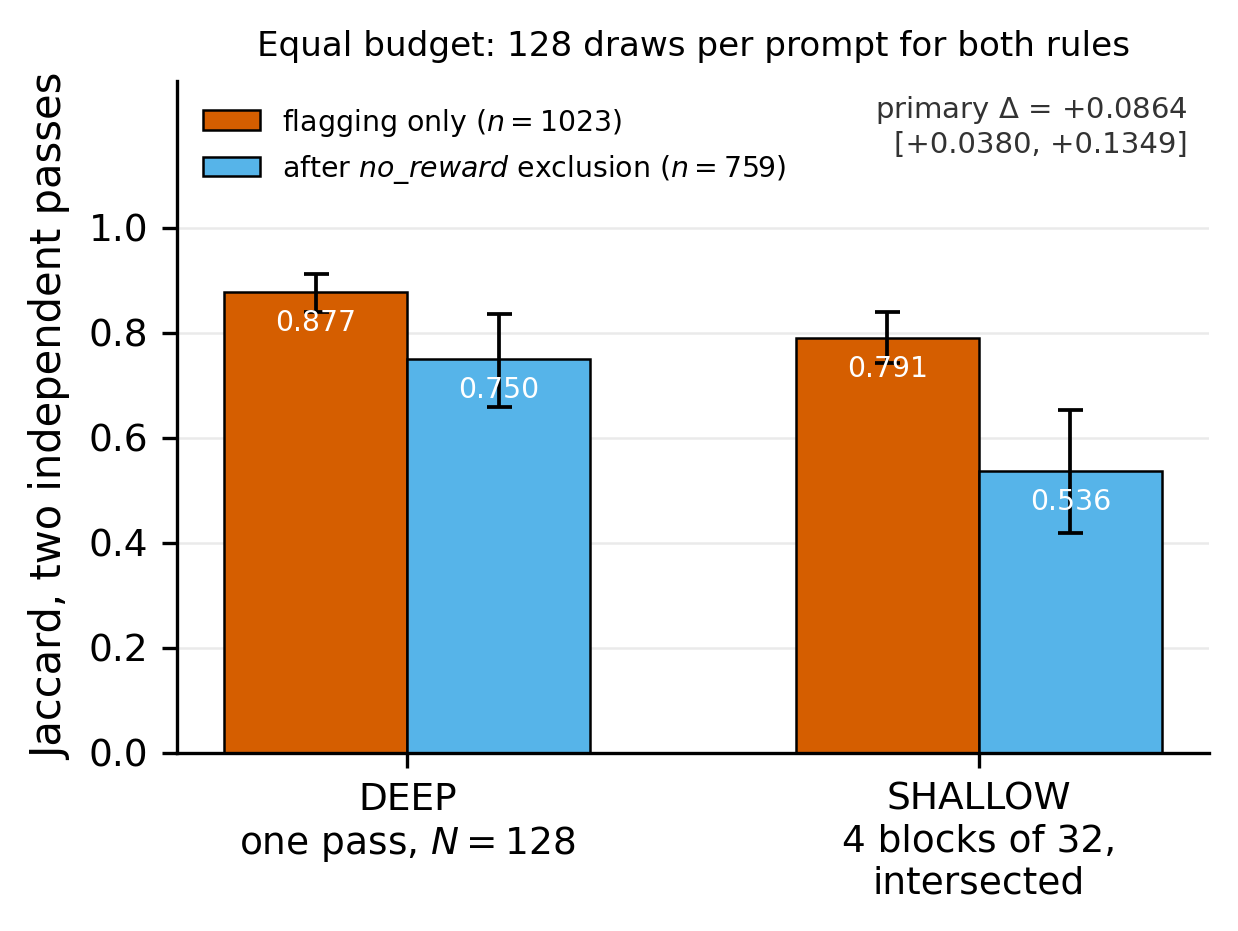}
  \caption{Equal-budget comparison of one deep \(N=128\) evaluation and four shallow blocks of 32. The deeper rule is more reproducible both before and after applying the never-rewarded exclusion. Error bars show 95\% intervals.}
  \label{fig:appendix-deep-shallow}
\end{figure}

Figure~\ref{fig:appendix-deep-shallow} shows the complete equal-budget comparison. This experiment isolates evaluation noise on one fixed checkpoint; it is not a full replication of the published multi-seed construction, which also includes training variation. The result therefore supports a narrower conclusion: when estimating a threshold-defined set from a fixed policy, splitting a fixed budget into several thresholded blocks is less reproducible than using the same observations in one deeper estimate.

\subsection{Validation of the sampling-budget model}
\label{app:validation}

Before making prospective predictions, we evaluated the model against several measurements that already existed for other parts of the study. The full retrospective and consistency results are shown in Table~\ref{tab:appendix-retrospective}. The table contains seven rows. Six of them, the four Qwen2.5-0.5B rules and the two Llama-3.2-3B passes, are treated as retrospective or cross-model consistency checks. The seventh, Qwen2.5-1.5B, is shown as an excluded diagnostic and is not counted as validation evidence, for the reasons given below.

\begin{table*}[t]
  \centering
  \small
  \setlength{\tabcolsep}{5pt}
  \renewcommand{\arraystretch}{1.12}
  \begin{tabularx}{\linewidth}{@{}
    >{\raggedright\arraybackslash}X
    >{\centering\arraybackslash}p{0.13\linewidth}
    >{\centering\arraybackslash}p{0.13\linewidth}
    >{\centering\arraybackslash}p{0.11\linewidth}@{}}
    \toprule
    \rowcolor{gray!18}
    \textbf{Model / rule} & \textbf{Measured} & \textbf{Predicted} & \textbf{Error} \\
    \midrule
    \rowcolor{gray!3}
    Qwen2.5-0.5B, one pass \(N=32\) & 0.7977 & 0.8000 & +0.0023 \\
    Qwen2.5-0.5B, one pass \(N=128\) & 0.8770 & 0.8759 & -0.0011 \\
    \rowcolor{gray!3}
    Qwen2.5-0.5B, four blocks of 32 & 0.7906 & 0.8228 & +0.0322 \\
    Qwen2.5-0.5B, two blocks of 16 & 0.7028 & 0.7248 & +0.0220 \\
    \rowcolor{gray!3}
    Llama-3.2-3B, two blocks of 16 & 0.6844 & 0.6763 & -0.0081 \\
    Llama-3.2-3B v2, two blocks of 16 & 0.7070 & 0.6926 & -0.0144 \\
    \rowcolor{gray!3}
    Qwen2.5-1.5B, two blocks of 16 & 0.6849 & 0.6718 & -0.0131 \\
    \bottomrule
  \end{tabularx}
  \caption{Retrospective and cross-model consistency checks for the sampling-budget model. Six rows are treated as retrospective or consistency checks; the Qwen2.5-1.5B row is an excluded diagnostic and is not counted as validation evidence. The cross-model rows reuse the generation pass from which each pass-rate distribution was estimated and are therefore consistency checks rather than independent validation results.}
  \label{tab:appendix-retrospective}
\end{table*}

The first two deep evaluations are predicted to within 0.002. The largest discrepancy occurs for the four-block rule, where the model predicts 0.8228 against an observed 0.7906. Importantly, the model still recovers the ordering between deep and shallow evaluation, although it understates the measured gap: the predicted difference is 0.0531, compared with the observed 0.0864. The discrepancy is not explained by incorrect predicted set sizes, which match the observed sizes to within 1.8\%, nor by obvious within-pass dependence: the observed block variance is 0.985 of the binomial expectation. We therefore retain the residual rather than fitting an additional correction to it. The cross-model rows should be read only as consistency checks. Each model's pass-rate distribution is estimated from the same \(N=32\) generation pass subsequently divided into two blocks, so these are not independent out-of-sample validations. The Qwen2.5-1.5B point is weaker still: 66.37\% of its prompts are flagged and 79.44\% of its rollouts are unparseable under our generation cap. In that setting, the measured set largely reflects the binding generation budget, and we do not count the point as validation evidence.

A simpler model could reproduce the observed number of flagged prompts by assigning every prompt a single common success probability. Such a model deliberately removes the heterogeneity in prompt difficulty while preserving the marginal flagging rate. If set reproducibility were determined mainly by the final set size, this null should predict repeated-set agreement reasonably well. It does not. Across the retrospective comparisons, the homogeneous model misses measured Jaccard by approximately 0.19--0.53. The distributional model misses the four block-based comparisons in Table~\ref{tab:appendix-retrospective} by only 0.008--0.032, and the two single-pass deep evaluations by less than 0.003. Matching the marginal number of flagged prompts is therefore insufficient. What matters is how the latent success probabilities are distributed around the threshold: two prompt collections can produce the same set size while having very different probabilities of assigning the same individual prompts on a repeat evaluation. This comparison is important for interpreting the budget calculation. The model is not simply learning how many prompts should be flagged at a given \(N\); it uses the shape of the pass-rate distribution to predict which membership decisions are likely to change.

\begin{figure}[t]
  \centering
  \includegraphics[width=0.65\linewidth]{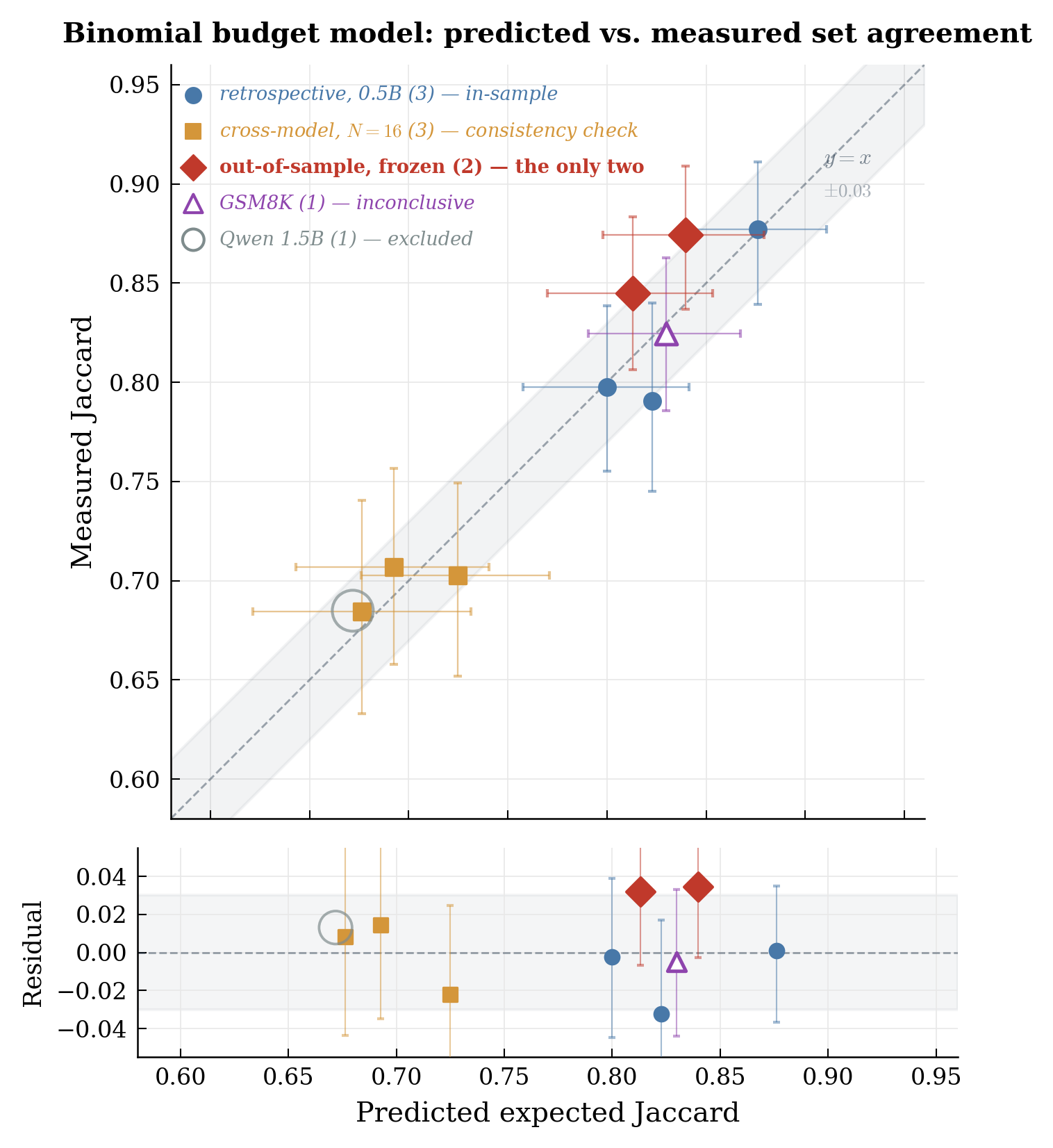}
  \caption{Predicted versus measured Jaccard agreement across retrospective checks, cross-model consistency checks, and prospectively frozen predictions. Filled diamonds mark the two prospective measurements; the Qwen2.5-1.5B point is shown separately and is not counted as validation evidence.}
  \label{fig:appendix-predicted-measured}
\end{figure}

The strongest validation in this study uses two sampling budgets that had not yet been measured. We froze and timestamped the predictions before generating the corresponding evaluation responses. The registered values were
\[
N=40:
\qquad
J_{40}=0.8131\,[0.7699,0.8533],
\]
and
\[
N=64:
\qquad
J_{64}=0.8397\,[0.7980,0.8792].
\]
Only after these predictions were fixed did we generate the new evaluations, totaling 212,784 responses. The measured Jaccard values were
\[
N=40:
\qquad
0.8450\,[0.8063,0.8833],
\]
and
\[
N=64:
\qquad
0.8742\,[0.8369,0.9091].
\]
Both measurements fall inside their frozen prediction intervals. At \(N=40\), the complete measured interval also lies above the target reproducibility of 0.80, so the practical budget recommendation survives the prospective test rather than merely the point prediction. Both prospective residuals have nearly the same sign and magnitude: the model underpredicts reproducibility by roughly 0.03. We do not remove this offset after observing it. One possible contributor is that repeated evaluations share a deterministic verifier, which can create positive dependence not represented by a fully independent binomial calculation. This explanation was anticipated, but the retrospective checks contain residuals in both directions, so the available evidence does not establish it as the cause. For the practical budget calculation, the observed direction is conservative: underprediction recommends slightly more sampling rather than less. Figure~\ref{fig:appendix-predicted-measured} shows all retrospective, consistency, and prospective points together. The two frozen prospective measurements are marked separately from the retrospective checks so that the figure does not visually overstate the amount of independent validation.

\subsection{Discrete thresholds and what transfers across settings}
\label{app:discrete-thresholds}

The dense budget curve contains a feature that would disappear if only conventional powers of two were reported. Over small budgets, predicted reproducibility frequently decreases when \(N\) increases by one. A concrete example is
\[
J_{10}=0.7417,
\qquad
J_{19}=0.7227,
\]
so almost twice as many responses can correspond to a lower predicted agreement. The difference itself is small relative to the uncertainty of either estimate; the important observation is the systematic sawtooth pattern rather than this individual pair. The source is the discrete decision boundary. At \(\tau=0.1\), \(N=10\) permits one success while remaining at the threshold, whereas nearby values of \(N\) can implement a smaller effective cutoff until the next integer boundary is reached. The estimator therefore becomes statistically sharper at the same time that the classification rule changes. Over \(N=8\) to \(160\), this effect produces frequent local declines; beyond that range the sawtooth becomes much less prominent. This is precisely why the paper defines a required budget as the point from which a target remains satisfied, rather than the first value of \(N\) at which the curve crosses it.

The same issue also limits comparisons between aggregation rules. At very small block sizes, \(\lfloor\tau B\rfloor=0\), so a shallow rule becomes equivalent to requiring zero correct responses in every block. It is then no longer an alternative estimator of the same threshold used by the deeper rule. For \(\tau=0.1\), no useful deep-versus-shallow comparison can be formed from the available 32-draw screening passes without changing the effective threshold. The equality at \(128\) versus \(4\times32\) is therefore a fortunate property of that particular comparison, not something that should be expected generally.

The numerical budget values in this paper should not be carried directly to another model or dataset. A model for which most prompts lie far from \(\tau\) can obtain stable labels with relatively few responses, while another model with substantial probability mass near the same threshold may require much deeper evaluation. Changing \(\tau\) can alter the required budget just as strongly. The same consideration applies during training as a model improves: a fixed number of responses need not provide the same reliability throughout the post-training trajectory. What does transfer is the procedure. Define the prompt-level quantity of interest, specify the threshold, estimate the distribution of latent success probabilities while accounting for finite-sample noise, and compute the predicted repeatability of the resulting set before treating membership as fixed. If the available budget cannot achieve the required reproducibility, the uncertainty belongs in the downstream analysis rather than being removed by an arbitrary aggregation rule.

\section{Gradient and downstream sensitivity analyses}
\label{app:gradient-sensitivity}

This appendix collects the analyses supporting Section~\ref{sec:gradient}. Figure~\ref{fig:matched_k} in the main paper remains the primary figure; the material here records the downstream propagation of set uncertainty and the cohort details behind the matched comparison.

\subsection{Propagation of set uncertainty to gradient similarity}
\label{app:propagation}

\begin{figure}[t]
  \centering
  \includegraphics[width=0.68\linewidth]{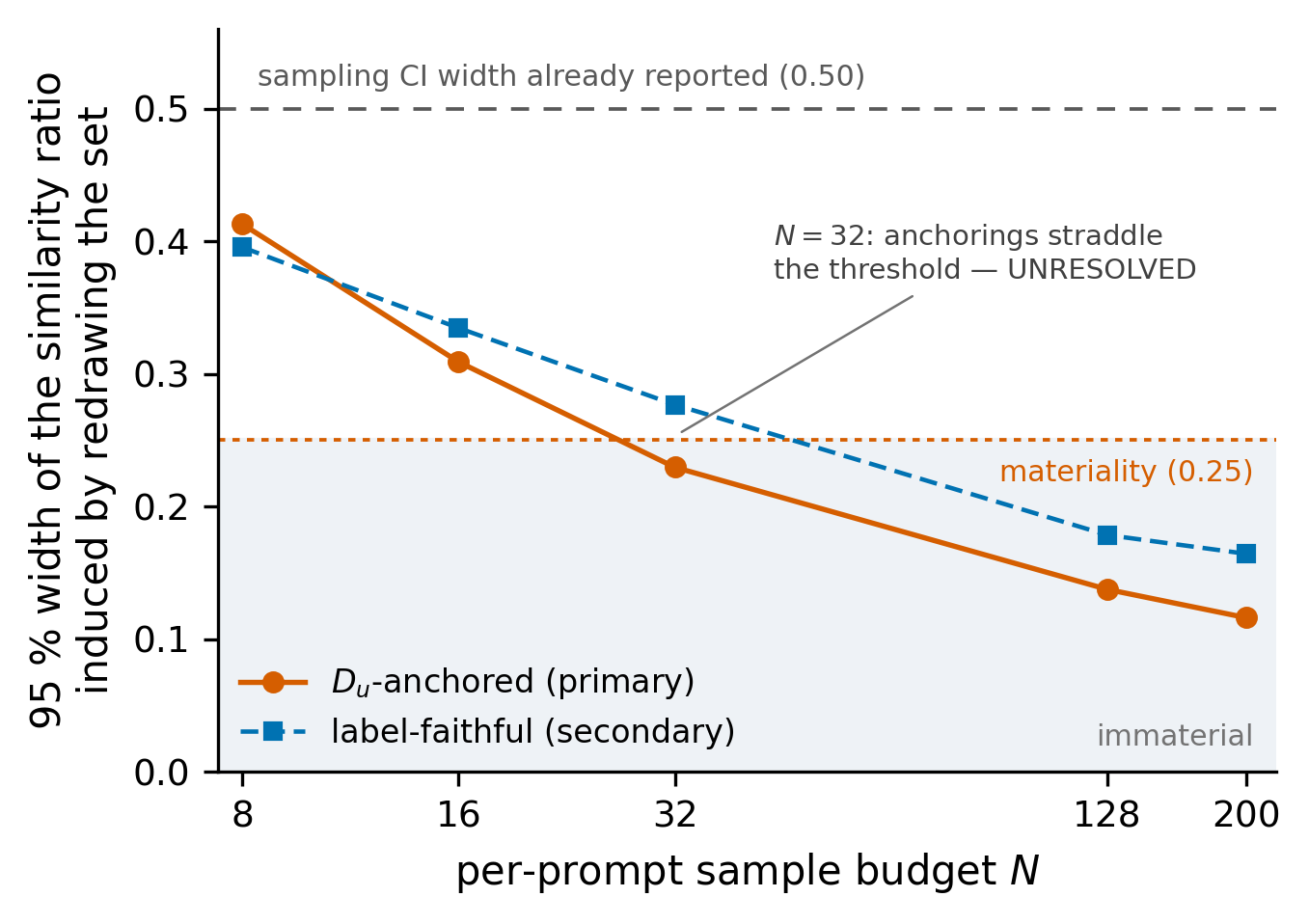}
  \caption{Full 95\% interval width of the easy-to-\(D_u\) gradient-similarity ratio induced by redrawing the threshold-defined set from the validated sampling model, as a function of per-prompt budget \(N\). The \(D_u\)-anchored variant is primary and the label-faithful variant secondary. The dotted line marks the pre-specified 0.25 materiality threshold; the dashed line marks the 0.50 sampling confidence-interval width already reported for the downstream statistic. At \(N=32\) the two anchorings straddle the threshold.}
  \label{fig:appendix-propagation}
\end{figure}

Low Jaccard is only scientifically important if uncertainty in membership is
large enough to affect something computed on the selected population. We
therefore redraw realizations of the threshold-defined set from the validated
sampling model and recompute the easy-to-\(D_u\) gradient-similarity ratio.
At \(N=8\), membership uncertainty alone contributes a full 95\% interval width
of \(0.4128\) to that ratio; at \(N=16\), the full width is \(0.3090\). These are
83\% and 62\%, respectively, of the full 95\% sampling confidence-interval width
already reported for the downstream statistic. At \(N=32\), the \(D_u\)-anchored
and label-faithful variants fall on opposite sides of our pre-specified 0.25
materiality threshold, so we mark that budget unresolved. By
\(N=128\), the additional full 95\% interval width falls to \(0.1375\), and at
\(N=200\) to \(0.1161\). Set uncertainty is therefore material at the
small assignment budgets commonly used for difficulty labels and becomes much
less important once the per-prompt measurement is sufficiently deep. Figure~\ref{fig:appendix-propagation} shows the full
budget-dependent propagation for both anchorings.

\subsection{Matched-gradient analysis and cohort composition}
\label{app:matched-gradient}

The matched comparison in Figure~\ref{fig:matched_k} is computed on a fixed cohort and repeated at \(K\in\{4,6,8,11\}\) correct rollouts averaged per prompt. At every tested level the easy-to-\(D_u\) similarity ratio lies below the unmatched value of 2.327 times and above one; the headline \(K=11\) comparison gives \(1.539\) times with a 95\% interval of \([1.26,1.88]\).

Matching introduces attrition. Requiring \(K=11\) correct rollouts leaves 28 of the 71 \(D_u\) prompts in the \(N=200\) gradient cohort eligible. The fixed matched cohort used in Figure~\ref{fig:matched_k} contains 28 \(D_u\), 28 learnable, and 38 easy prompts. The retained \(D_u\) prompts are necessarily those that produced more successful responses, so they are easier than many of the prompts removed by the matching requirement.

This selection also explains the difference between two sample counts reported in the paper, which refer to different cohorts. Across the full \(N=200\) gradient cohort, \(D_u\) prompts have a median of 8 correct rollouts, compared with 65 for easy prompts. Within the fixed matched cohort, the natural, unmatched counts shown by the open markers in Figure~\ref{fig:matched_k} are \(K=16\) for \(D_u\), \(K=18\) for learnable, and \(K=64\) for easy prompts; the unmatched 2.327 times ratio compares easy prompts at \(K=64\) with \(D_u\) prompts at \(K=16\).

On a log scale, matching removes 48.9\% of the observed separation, with an interval of \([29.3,71.2]\%\). We do not interpret this as a causal decomposition, because matching changes cohort membership: the matched ratio is evidence that unequal sample count contributes substantially to the original gradient gap, not a corrected estimate for the full \(D_u\) set. The evaluation-side generation-length control for set membership is reported in Appendix~\ref{app:generation-length}.

\section{Optimization diagnostics and training impact}
\label{app:optimization-diagnostics}

This appendix collects the diagnostics supporting the optimization analyses in Section~\ref{sec:gradient} and the training experiment in Section~\ref{sec:training_effect}.

\subsection{Exposure, zero-variance filtering, and GRPO normalization}
\label{app:exposure-normalization}

With \(G=8\), a prompt contributes no GRPO update on a step where all sampled rewards are identical. \(D_u\) prompts receive an average of 2.14 contributing steps per run, compared with 3.13 for learnable prompts and 5.26 for easy prompts. None of the 156 prompts in the exposure-analysis cohort is completely starved across all five runs, and 35 of them, or 22.4\%, receive at least the mean exposure of the learnable group while still improving slowly. This rules out the simple form of exposure starvation in which \(D_u\) prompts fail to learn because they receive no usable update opportunities. It does not rule out content starvation, because the training-time response strings were not stored.

The exposure counts above are counts of steps that pass GRPO's zero-variance filter. Under the GRPO sampling procedure, a prompt group whose sampled rewards have zero within-group variance is discarded from the update, so a prompt whose eight rollouts are all correct or all incorrect on a given step contributes nothing on that step. Prompts that become consistently solved or consistently failed therefore disappear from subsequent sampling, which is why the paper characterizes cohorts by their reward slope rather than by terminal reward (Appendix~\ref{app:slow-learning}).

A separate candidate explanation is that GRPO's reward-standard-deviation normalization suppresses updates on hard prompts on the steps where they do contribute. Conditional on a prompt producing non-zero reward variance, the mean \(1/\sigma\) scale is 2.8360 for \(D_u\) prompts and 2.4262 for easy prompts, about 17\% larger for the harder group. A reduced normalization scale is therefore not supported as the explanation; the main asymmetry is in how often a usable update is formed.

\subsection{Does difficulty construction materially change training?}
\label{app:training-impact}

The training experiment uses two neighboring difficulty constructions that remove very different amounts of usable training signal. After accounting for GRPO's zero-variance filter, one construction removes approximately 5.0\% of the usable training signal and the other 27.6\%, a difference of more than fivefold.

We train five seeds under each construction and compare them with size-matched random-removal controls, evaluating on held-out accuracy.

Despite the large difference in removed signal, the resulting models differ by less than one held-out accuracy point. The registered sensitivity of this experiment is not sufficient to resolve effects of that size, so this result is not evidence of equivalence. It does, however, establish a boundary on the paper's claim: an unstable difficulty definition can strongly change the reported prompt set without necessarily producing an equally large change in downstream training performance.

\section{Experimental configuration and replication audit}
\label{app:replication-audit}

\subsection{Experimental configuration and statistical procedures}
\label{app:replication-audit-config}

Our primary experiments use Qwen2.5-0.5B on the 1,023-prompt MATH training subset used throughout the study. RLVR training follows the GRPO setup of \citet{chen2026unlearnability}, with \(G=8\) sampled responses per prompt and five independent training seeds. We use the original answer verifier without modification. Set-membership analyses are performed primarily with \(N=128\) evaluation rollouts per prompt, while the gradient analyses use \(N=200\) rollouts from the initial policy.

\paragraph{Existing assets.}
We use Qwen2.5-0.5B under the Apache 2.0 license, the MATH dataset under the MIT license, and the original unlearnability RLVR implementation under the Apache 2.0 license. All external assets are cited in the paper and used in accordance with their stated licenses and terms.

The experiments in this paper are a replication and extension rather than a byte-for-byte rerun of the original implementation. We therefore audited the effective configuration from the actual training and evaluation execution paths. The audit covers every shell variable passed to training and every data or sampling option forwarded to evaluation. At commit \texttt{1dce7fe}, we identified 47 effective constants: 30 match the original configuration and 17 differ.

\begin{table*}[ht]
  \centering
  \footnotesize
  \setlength{\tabcolsep}{6pt}
  \renewcommand{\arraystretch}{1.14}
  \arrayrulecolor{gray!60}
  \begin{tabularx}{\linewidth}{@{}
    >{\raggedright\arraybackslash}p{0.34\linewidth}
    >{\raggedright\arraybackslash}X
    >{\raggedright\arraybackslash}X@{}}
    \toprule
    \rowcolor{gray!18}
    \textbf{Configuration constant} & \textbf{Ours} & \textbf{Original} \\
    \midrule
    \rowcolor{gray!10}
    \multicolumn{3}{@{}l}{\textbf{Training}} \\
    \rowcolor{gray!3}
    Dataset name & \path{simplelr_qwen_level1to4_sub1k} & \path{simplelr_qwen_level1to4} \\
    Maximum response length & 1,024 & 5,120 \\
    \rowcolor{gray!3}
    Train batch size & 32 & 256 \\
    PPO mini-batch size & 16 & 64 \\
    \rowcolor{gray!3}
    PPO micro-batch size & 2 & 32 \\
    Log-probability micro-batch size & 4 & 128 \\
    \rowcolor{gray!3}
    Micro rollout batch size & 128 & 1,024 \\
    Tensor parallelism & 1 & 2 \\
    \rowcolor{gray!3}
    GPUs per node & 1 & 4 \\
    Rollout GPU memory utilization & 0.45 & 0.75 \\
    \rowcolor{gray!3}
    Total epochs & 12 & 50 \\
    Save frequency & 40 & 20 \\
    \rowcolor{gray!3}
    Test frequency & 100,000 & 5 \\
    Validation batch size & 200 & 1,000 \\
    \addlinespace[2pt]
    \rowcolor{gray!10}
    \multicolumn{3}{@{}l}{\textbf{Evaluation}} \\
    \rowcolor{gray!3}
    Test data & Training subset used here & Original test split \\
    Maximum generation length & 1,024 & 5,120 \\
    \rowcolor{gray!3}
    GPU memory utilization & 0.60 & 0.75 \\
    \bottomrule
  \end{tabularx}
  \caption{Configuration differences identified from the executed training and
  evaluation paths. The main text highlights differences that directly affect
  scientific interpretation; this table records the complete audit.}
  \label{tab:config-audit}
\end{table*}

We divide the differences into four classes. \textbf{Direct-science} differences change the data or generation budget and can directly alter the measured quantities. \textbf{Optimization} differences modify the update protocol. \textbf{Execution-not-invariant} differences arise from adapting the original multi-GPU setup to a single-GPU execution path; we do not assume these changes are bitwise or stochastically invariant. \textbf{Operational} differences affect logging, checkpointing, or evaluation frequency but do not directly define the training objective. Table~\ref{tab:config-audit} lists the complete set of differing constants; it does not annotate each row with its class.

The most important difference for the results in this paper is the generation budget. Our training and primary evaluations use a maximum response length of 1,024 tokens, whereas the original scripts default to 5,120. This difference affects hard prompts more strongly than easy prompts and therefore cannot be treated as an ordinary implementation detail. We address the evaluation-side consequence directly by repeating the full \(N=128\) evaluation with a 5,120-token budget. The training runs themselves remain at 1,024 tokens, so none of our training-dependent results should be interpreted as an exact numerical reproduction of the original configuration.

The dataset also differs from the original execution path: our experiments use the 1,023-prompt subset analyzed throughout this paper rather than the larger dataset referenced by the original baseline script. This choice fixes the set of prompts across the replication, deep evaluations, sampling-budget analysis, and gradient experiments. It also means that numerical quantities such as set size and required sampling budget are specific to this experimental setting.

\paragraph{Compute resources.}
All reported GPU experiments were run on a single NVIDIA GeForce RTX 3090 Ti with 24 GB of VRAM. Most training and evaluation runs took approximately 16--24 hours to complete, while lightweight post-processing and statistical analyses required substantially less compute.

Confidence intervals throughout the paper are obtained by bootstrap, with prompts resampled at the level appropriate to each statistic, and are reported as 95\% intervals in square brackets. Reward slopes are reported per 100 optimizer steps. The two prospective budget predictions in Appendix~\ref{app:validation} were frozen and timestamped, together with their intervals, before the corresponding evaluation responses were generated.

\subsection{Generation length and answer extraction}
\label{app:generation-length}

The 1,024-token cap produces a substantial number of truncated responses. On the reference evaluation, increasing the budget from 1,024 to 5,120 tokens reduces the overall truncation rate from \(0.0961\) to \(0.0320\), and the answer-extraction failure rate from \(0.0977\) to \(0.0329\). Almost all extraction failures at the shorter budget are truncations rather than failures of the verifier itself.

Crucially, this correction does not substantially change the difficulty-defined set. The \(N=128\) sets obtained at 1,024 and 5,120 tokens have Jaccard similarity \(0.8984\,[0.8639,0.9310]\). Two independent evaluations at the same budget agree at \(0.8770\,[0.8397,0.9127]\). The movement produced by increasing the token budget is therefore no larger than the repeatability variation already present when the evaluation configuration is held fixed.

This control addresses only the evaluation-side set-membership result. Because our training trajectories were generated with the shorter limit, it does not establish that the training curves or gradient quantities would be numerically unchanged under the original 5,120-token configuration.

\end{document}